# Generalized Scene Reconstruction

John K. Leffingwell, Donald J. Meagher, Khan W. Mahmud and Scott Ackerson

Quidient, LLC, Columbia, MD

*Abstract*—The feasibility of new passive approach to 3D reconstruction, called Generalized Scene Reconstruction (GSR), is explored in this paper. Generalized scenes are defined to be "boundless" spaces that include non-Lambertian, partially transmissive, textureless and finely-structured matter. GSR enables such scenes to be effectively reconstructed by devices-using-scene-reconstruction (DSRs) such as mobile phones, augmented reality (AR) glasses and drones. A new data structure called a plenoptic octree is introduced to enable efficient (database-like) light and matter field reconstruction in DSRs. To satisfy threshold requirements for reconstruction accuracy, scenes are represented as systems of partially polarized light interacting with matter. To demonstrate GSR, a prototype imaging polarimeter is used to reconstruct, by sensing and modeling generalized polarimetric light fields, highly reflective, hail-damaged automobile body panels. Follow-on GSR experiments are described.

## I. Introduction

Scene Reconstruction Engines (SREs) create 3D scene models from digital images using a process called scene reconstruction. SREs are critical, new-category components of devices-using-scene-reconstruction (DSRs) such as 3D mobile phones, augmented reality (AR) glasses and drones. Generalized Scenes are "boundless" 3D spaces, full of electromagnetic energy (light) and overlapping matter that includes highly non-Lambertian, partially transmissive, textureless and finely-structured matter, like matter present in contemporary offices, flower gardens and modern kitchens (see Fig. 1).

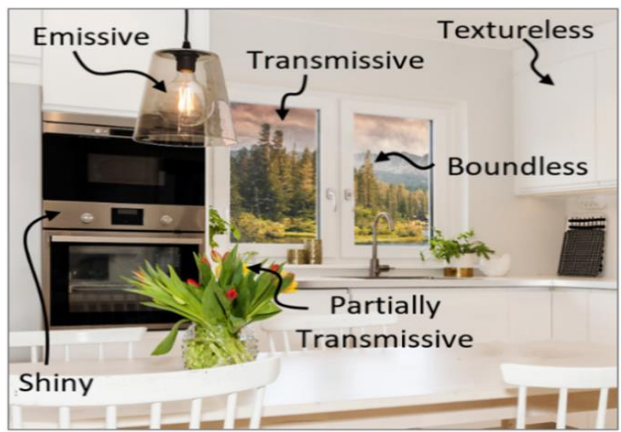

Fig. 1.  Generalized Scene

To enable emerging applications such as AR, 3D photography, remote medicine, mass customization (e.g., aim DSR at window to order fitted curtains) and visual analytics (e.g., aim DSR at dented car panel to schedule repair), DSRs must be capable of generalized scene reconstruction (GSR). After all, many of the spaces that people occupy in the course of their daily personal and professional lives are "generalized" in the sense of the generalized scene definition above. But, today's SREs are unable to satisfy broad requirements for generalized scene reconstruction, which requirements can be stated in terms including accuracy, size, power and affordability requirements. Referring to Fig. 1, the highly reflective countertops, the partially transparent lamp, the textureless white cabinets, the plants and the outdoor scene visible through the windows cannot be accurately reconstructed by DSRs available today. A severe SRE price/performance bottleneck exists.

### A. Related Work

Passive scene reconstruction is an important problem in computer vision. Reconstruction of scenes comprised of sparse features is a well-studied problem [[31], [37], [38], [39]]. Reconstruction of scenes comprised of surfaces, often referred to as dense reconstruction, is an active area of research [[10], [53], [62], [63], [64], [65]]. The use of polarimetric characteristics of light to accomplish dense object reconstruction using imaging polarimeters is a recent topic of research [[2], [66]].

The technologies that are adapted and merged to implement GSR are well-studied: shape-from-polarization [[7], [58], [59], [60], [61]], imaging polarimetry [[41], [42], [43], [44], [45], [46]] spatial subdivision [[54], [55], [56], [57]], light fields [[68], [69], [70]] and light transport [[47], [48], [49], [50], [51], [52], [67]]. There are currently few references to the GSR approach itself [[1], [13], [27]].

### B. Contributions of This Paper

Contributions of this paper include: i) The introduction of a new hierarchical, spatio-directionally sorted data structure called a plenoptic octree to represent the light/matter field that exists in a generalized scene, ii) The introduction of a new projection method to efficiently compute light transport in a generalized scene, and iii) The extension/blending of the concepts of multi-view stereo (MVS) and shape-from-X (SfX), including shape-from-polarization (SfP), to GSR in ways that include: use of Bidirectional Light Interaction Functions (BLIFs) to represent the potentially omni-directional interaction of light with homogeneous and heterogeneous media, and use of potentially disjoint surface elements (surfels) to separate regions within media elements (mediels).

## II. IMPLEMENTATION

Our implementation brings together three primary building blocks to accomplish GSR: light field physics, scene learning, and spatial processing.

### A. Light Field Physics

In our GSR implementation, a light field physics module models the interaction between volumetric media elements ("mediels") and the radiometric light field elements ("radiels") that enter and exit the mediels. We define media as a volumetric region that includes some or no matter and in which light flows. Media can be homogeneous (e.g., empty space, air and water) or heterogeneous (e.g., the surface of a pane of glass, and the branch of a pine tree). Such interactions are complex in real scenes when a large number of known phenomena are included. Our light field physics module uses a simplified "light transport" model to represent these interactions. This is diagrammed in Fig. 2.

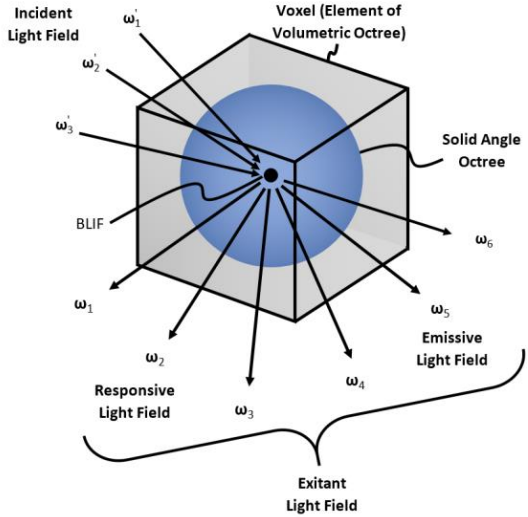

Fig. 2. Light field at a mediel

Our light transport model is mathematically expressed in Eq. (1) relating the incident and exitant light fields at a mediel. The first term represents the emissive light field emitted by the mediel when not stimulated by incident light. The responsive light field, represented by the second term, is determined by the incident radiels and the bidirectional light interaction function ("BLIF"), denoted $f_\ell(\mathbf{x} \to \boldsymbol{\omega}, \mathbf{x}' \leftarrow \boldsymbol{\omega}')$. We largely follow the notations of [12] in describing the essential light field relations at a mediel.

$$L(\mathbf{x} \to \boldsymbol{\omega}) = L_e(\mathbf{x} \to \boldsymbol{\omega}) + \int_{X'} \int_{\Omega'_{4\pi}} f_\ell(\mathbf{x} \to \boldsymbol{\omega}, \mathbf{x}' \leftarrow \boldsymbol{\omega}') L(\mathbf{x}' \leftarrow \boldsymbol{\omega}') \, d\boldsymbol{\omega}' \, d\mathbf{x}'$$

(1)

where $\mathbf{x}$ is the (position of the) mediel under consideration, $\mathbf{x}'$ is a mediel that contributes to the radiance exitant at $\mathbf{x}$, $X'$ is all mediels that contribute to the radiance exitant at $\mathbf{x}$ ($X' = \mathbf{x}$ in the case of no "light hopping", e.g., subsurface scattering), $\mathbf{x} \to \boldsymbol{\omega}$ is the radiel exitant in direction $\boldsymbol{\omega}$ at mediel $\mathbf{x}$, $\mathbf{x}' \leftarrow \boldsymbol{\omega}'$ is the radiel incident from direction $\boldsymbol{\omega}'$ at mediel $\mathbf{x}'$, $L(radiel)$ is the radiance of the indicated radiel, $L_e$ is emissive radiance, $f_\ell(\mathbf{x} \to \boldsymbol{\omega}, \mathbf{x}' \leftarrow \boldsymbol{\omega}')$ is the BLIF (with light hopping) yielding the radiance of exitant radiel $\mathbf{x} \to \boldsymbol{\omega}$ as a function of the radiance of incident radiel $\mathbf{x}' \leftarrow \boldsymbol{\omega}'$, $d\boldsymbol{\omega}'$ is the (amount of) solid angle subtended by radiel $\mathbf{x}' \leftarrow \boldsymbol{\omega}'$, $d\mathbf{x}'$ is the (amount of) surface area represented by mediel $\mathbf{x}'$, and $\Omega'_{4\pi}$ is the entire sphere ($4\pi$ steradians) of incident radiels. Eq. (1) may be straightforwardly extended to include time, polarization state, and wavelength as parameters.

When light is transported through empty space (e.g., dry air, which may be modeled as empty in many applications), radiance is conserved along the path of propagation. A mediel of empty space has a BLIF that is the identity function: the exitant light field is equal to the incident light field in all radiels. In a scene model consisting of non-empty media regions in empty space, conservation of radiance is used to transport light between the non-empty mediels.

A mediel can serve to represent a spatial element of a surface, where we define "surface" to mean the objective, resolution-dependent boundary between regions of dissimilar media type. At resolutions where a represented surface element has low local curvature, we call the mediel a "surfel", and a surface may be represented as a collection of surfels. When the media region on one side of a surfel is opaque (non-transmissive), the BLIF reduces to a bidirectional reflectance distribution function (BRDF):

$$L(\mathbf{x} \to \boldsymbol{\omega}) = L_e(\mathbf{x} \to \boldsymbol{\omega}) + \int_{\Omega'_{2\pi}} f_r(\mathbf{x}, \boldsymbol{\omega} \leftarrow \boldsymbol{\omega}') L(\mathbf{x} \leftarrow \boldsymbol{\omega}') (\mathbf{n} \cdot \boldsymbol{\omega}') \, d\boldsymbol{\omega}'$$

(2)

where $f_r(\mathbf{x}, \boldsymbol{\omega} \leftarrow \boldsymbol{\omega}')$ is the BRDF relating the radiances of the exitant radiel $\mathbf{x} \to \boldsymbol{\omega}$ and the incident radiel $\mathbf{x} \leftarrow \boldsymbol{\omega}'$, $\mathbf{n}$ is the surface normal vector at surfel $\mathbf{x}$, $(\mathbf{n} \cdot \boldsymbol{\omega}')$ is a cosine foreshortening factor that balances its reciprocal present in the canonical BRDF definition, and $\Omega'_{2\pi}$ is the continuous hemisphere ($2\pi$ steradians) of incident radiels centered about $\mathbf{n}$. A surfel's BRDF is generally based on a microfacet distribution model in conjunction with the Fresnel equations for reflection of s-polarized and p-polarized light [[32], [33], [34], [35], [36]]. When transmissive media exist on both side of a surfel, a bidirectional transmittance distribution function (BTDF) term is added to the BRDF term to yield the total BLIF.

Our implementation optionally models the changing polarization state of light as it interacts with surfaces and propagates through transmissive media. When operating in polarimetric mode, a Stokes vector $\mathbf{S} = [S_0, S_1, S_2, S_3]$ replaces the scalar radiance $L$ in the preceding equations. The $S_3$ component is omitted when circular polarization is not modeled. Model-driven predictions of light field radiels, in conjunction with polarimetric observations of a scene's light field, enable the accurate reconstruction of reflective surfaces that are featureless in the traditional sense, i.e., surfaces lacking localized features that would be found by a feature detection algorithm such as SIFT. The optical physics details of modeling light propagation and interaction with mediels are given in more detail in a patent application [13].

**B. Scene Learning**

Given the light transport model of Sec. II.A, the light field (collection of radiels) and matter field (collection of mediels) may be reconstructed in a region of scene space. For a single mediel, Eq. 3 represents the mathematical optimization problem to be solved:

$$\underset{\substack{f_\ell(\mathbf{x}\to\boldsymbol{\omega}, X'\leftarrow\Omega'_{4\pi}),\\ L(X'\leftarrow\Omega'_{4\pi})}}{argmin} \sum_{\substack{\text{observed}\\ \mathbf{x}\to\boldsymbol{\omega}}} error\Big(L_{\text{observed}}(\mathbf{x}\to\boldsymbol{\omega}) - L_{\text{predicted}}\big(\mathbf{x}\to\boldsymbol{\omega}, f_\ell, L(X'\leftarrow\Omega'_{4\pi})\big)\Big)$$

(3)

where $f_\ell(\mathbf{x}\to\boldsymbol{\omega}, X'\leftarrow\Omega'_{4\pi})$ is the BLIF (with light hopping) yielding the radiance of exitant radiel $\mathbf{x}\to\boldsymbol{\omega}$ as a function of the radiances of the contributing incident radiels $X'\leftarrow\Omega'_{4\pi}$, $L(X'\leftarrow\Omega'_{4\pi})$ is the (set of) radiances of the contributing incident radiels, observed $\mathbf{x}\to\boldsymbol{\omega}$ is the set of observed radiels exiting voxel $\mathbf{x}$, $L_{\text{observed}}(\mathbf{x}\to\boldsymbol{\omega})$ is the radiance recorded by a camera that senses radiel $\mathbf{x}\to\boldsymbol{\omega}$, $L_{\text{predicted}}(\mathbf{x}\to\boldsymbol{\omega}, f_\ell, L(X'\leftarrow\Omega'_{4\pi}))$ is the radiance of exitant radiel $\mathbf{x}\to\boldsymbol{\omega}$ predicted by BLIF $f_\ell$ and incident light field $L(X'\leftarrow\Omega'_{4\pi})$, and $error(L_{\text{observed}} - L_{\text{predicted}})$ is a function (including robustification and regularization mechanisms) that yields an inverse consistency measure between observed and predicted radiels. An uncertainty-based weighting factor is applied to the difference (residual) between the observed and predicted radiance.

Eq. (3) is generally a multidimensional global optimization problem performed over the degrees of freedom of the BLIF and the light field. For a region of scene space comprising multiple mediels, as is typical when reconstructing an "object of interest", the optimization is performed over a set X of mediels. To make the mathematical optimization problem of Eq. (3) tractable in practice, our implementation proceeds in stages, holding some subset of the mediels' BLIF parameters and/or the radiels' radiometric parameters constant. This may roughly be thought of as the light field and matter field undergoing alternating refinement iterations. Operation is driven toward a reconstruction goal that is typically a desired accuracy at some desired spatial resolution in the scene region of interest.

The core GSR optimization problem expressed in Eq. (3) can be said to unify and subsume certain existing shape-from-X ("SfX") approaches, e.g., shape-from-polarization, shape-from-shading, structure-from-motion, shape-from-silhouette, etc. A broad spectrum of machine learning approaches is applicable in solving the unified problem, from traditional gradient descent algorithms to neural networks and reinforcement learning.

When starting a reconstruction problem, we initialize the BLIF parameters of mediels postulated to be non-empty. This is done using a combination of a priori scene information and SfX techniques (e.g., sparse structure-from-motion based on bundle adjustment). Traditional SIFT-like localized features serve to constrain the voxel regions where we expend computational energy on mediel BLIF solving. Following initialization, the alternating refinement cycle mentioned above commences between the light field and matter field. Note that the camera viewpoints may be further refined by freeing their pose parameters in iterations of the cost function in Eq. (3).

Given a continuous surface region imaged at sufficient resolution, our GSR approach can reconstruct its detailed shape at sub-millimeter precision. The present-day implementation that accomplishes this on dented automobile panels realizes Eq. 3 in a pipeline of pragmatic steps (shown in Figure 3):

1. Image the surface and surrounding scene at high dynamic range.
2. Reconstruct the camera and optical target ("tag") poses.
3. Reconstruct the estimated nominal (undented) surface using the tag poses. This becomes the prior model of the surface.
4. Reconstruct the incident light field at mediels lying on the prior surface.
5. Reconstruct the BRDF of the surface.
6. Reconstruct the normal vector of each surfel at 0.2° angular resolution, and then Fourier-integrate the normal vectors to reconstruct the dented surface deviations at 5μm depth resolution.

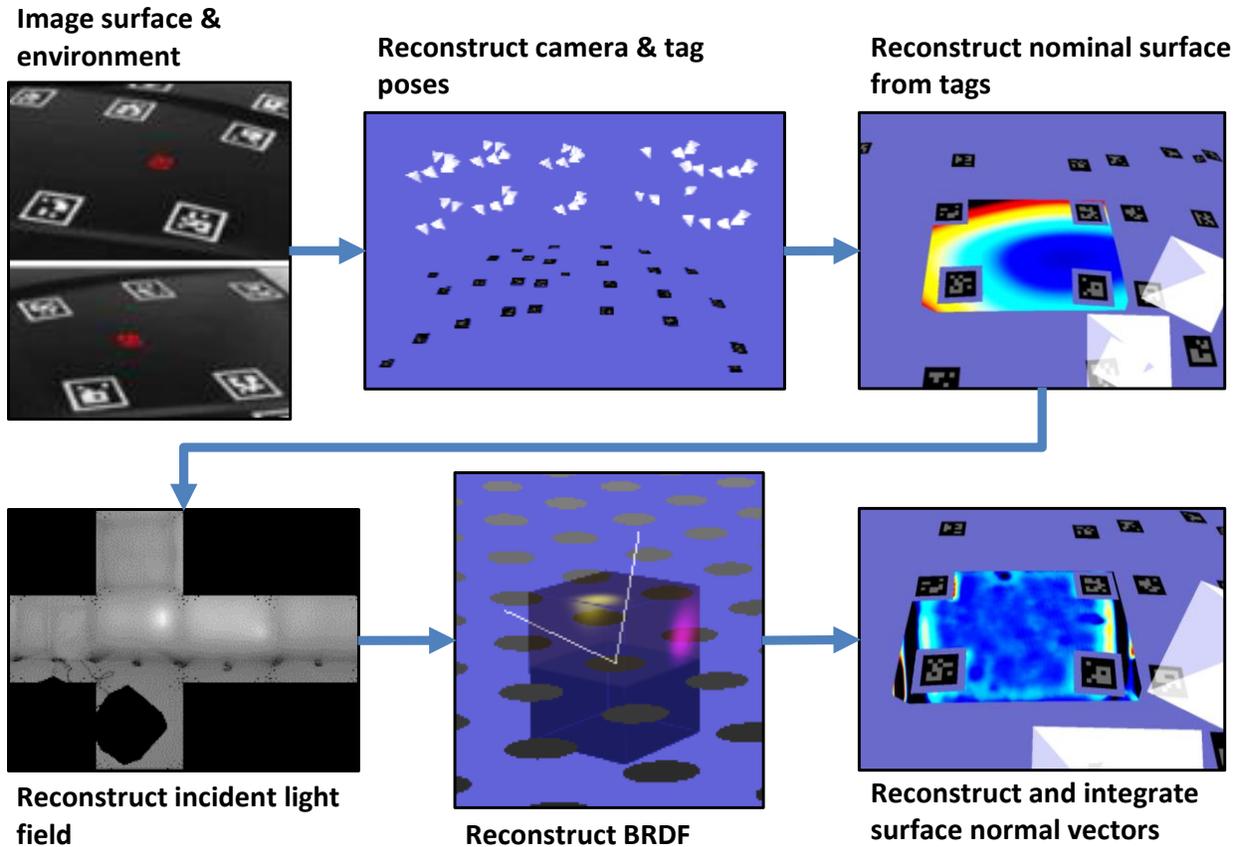

Fig. 3. Steps in regularized reconstruction of the sub-millimeter surface profile of a dented automobile panel

## C. Spatial Representation and Processing

Using the methods presented here, GSRs are being developed for use within inexpensive, mobile DSRs to implement a new generation of advanced applications for a wide range of uses. They will be used to acquire highly-accurate models of quotidian scenes that contain challenging materials subject to real-world lighting. Detailed light-field information will be acquired. In addition to modeling light's interaction with matter for use in reconstruction, it will facilitate the interpretation and understanding of the scene. It can also be used to realistically illuminate synthetic objects to blend naturally into the real-world, a problem in AR.

DSRs will support advanced modeling, recognition, 3D design, simulation and many other capabilities. For example, recognized objects can be replaced by preexisting models that have realistic characteristics and behaviors for use in accurate, physics-based interactions. It is expected that AI and Machine Learning will be more effective when analyzing rich, fully-evaluated, multi-dimensional representations of the real world, rather than images. Also, multiple users can employ powerful 3D CAD tools for distributed cooperative design. To maintain the perception of reality, high update rates (e.g., 90 Hz.) are needed from local computing resources (even if the system is connected to the Cloud).

An analysis was undertaken of the spatial representation and processing requirements needed in a Spatial-Processing Unit (SPU) to achieve these ambitious goals. First, a robust and unified representation is needed for all core processing operations. In addition, the SPU must support solid-modeling capabilities such as parametric modeling, set operations (union, intersection, etc.), mass properties (volume, weight, surface area, center of mass, etc.), connectivity analysis and others. And to meet the real-time image-generation requirements, direct rendering of the models is necessary (eliminating the costly extraction of surfaces for display). A major challenge was to integrate the modeling of light and its interaction with real and synthetic matter within this new modeling framework.

Perhaps the most difficult requirement was the need to handle unlimited real-world spatial data. The core limitation with existing methods is the computational growth as spatial datasets grow. A new method was needed to efficiently and effectively deal with "Big 3D" (the three-dimensional equivalent of Big Data).

**1) Innovations.** Existing modeling and visualization algorithms on CPUs and GPUs were found to be too inefficient for the core functions needed for DSRs. They were rejected. The power required for the billions or tens-of-billions of transistors currently precludes effective, real-time mobile use in small packages for quotidian scenes. To solve the problem, a radical departure from current methods was derived from the ground up. It was inspired by the development of raster graphics. At the start of the '80s, computer displays drew line drawings for viewing by sweeping an electron beam across the face of a CRT. As the number of lines increased, the hardware to compute the mathematics and sweep the beam in real time became expensive and limited display complexity.

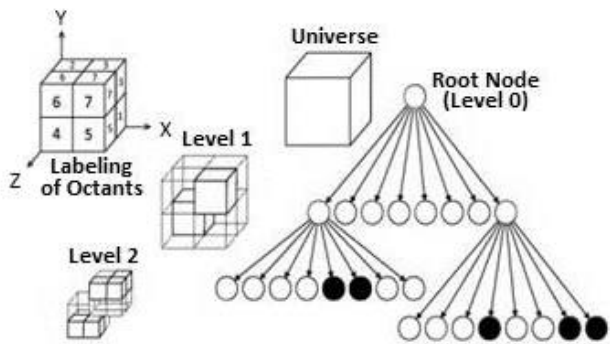

Fig. 4. Octree Data Structure

The solution was to quantize the display screen into an array of areas (picture elements or "pixels"). An unlimited number of lines could be written into the array and continuously refreshed to the viewer at a fixed rate, regardless of complexity. Plus the computational burden of quickly and continuously computing vector endpoints was removed. The hardware was often employed to write lines into the display, but only when the dataset or view changed (real-time operation is not necessary). As an added bonus, the dimensionality of the displayed elements increased, from 1D lines to 2D areas enabling the effective display of surfaces and images (real or synthetic).

The new method developed here employs major innovations in three areas:

a) Representation – In general, more robust representations enable more powerful algorithms, offering increased affordance. For example, spatial elements of higher dimensionality represent ranges of lower-dimension elements (e.g., a 1D line represents a range of 0D points). Thus, instead of employing surfaces and rays, both matter and light (and their interactions) are quantized and represented to a higher dimensionality (volumes for matter and solid angles for light).

b) Algorithms – Efficient algorithms were derived to exploit the characteristics of the new representations and data structures. This resulted in slow computational growth with scene complexity.

c) Implementation – A novel computing architecture was developed to implement the new method in software or hardware. The primary consideration for a hardware architecture was power consumption. The strategy was to exploit the simplified math and parallelism made possible by hierarchical, higher-dimensional quantization. The architecture is designed for massively parallel operation on simple computing elements with reduced clock rates.

2) **Representation.** Volumetric methods were adopted for this effort. The octree was adopted as the fundamental data structure for representing matter. Octree technology [[11], [14],[30]] was originally developed to reduce the computational growth of spatial processing algorithms, simplify computations and to facilitate parallelization. As shown in Fig. 4, an octree uses voxels (volume elements) at multiple levels of spatial resolution, beginning with one cube at level 0 representing the universe. The storage required is related to the surface area of the represented object or material (O(area/resolution2)), not the volume of the universe (as with voxel arrays).

Octrees, like triangles for surfaces, are a universal least-common-denominator spatial representation that can represent nearly any arrangement of scene media up to and including space-filling solids. They can be readily generated from points, lines, surfaces, solid shapes, volumetric data and so on. The core computation is to determine the other representation's spatial status relative to regularly subdividing cubical volumes in space (nodes in the octree). The node status can be one of the following: completely disjoint, completely occupied or other. And, just as line-drawing processors are used in raster display systems whenever the lines change, existing polygon display processors can be used to quickly convert new polygon datasets or when they change. They are transformed into the octree domain where they can be efficiently processed and displayed in logarithmic time. In many cases the conversion can be performed in the Cloud. Communications efficiency is enhanced in that only the nodes at the spatial resolutions needed are sent to the local display device (depending on the distance to the viewpoint).

Volumetric methods have been extensively used for medical visualization and surgical planning [16], plus specialized uses such as 3D shape matching, for some time. Octree methods are increasingly being used to process real-world spatial information, and octree methods are being applied, even with GPUs, to increase efficiency [[17], [18], [19]]. Ref. [20], for example, is implemented using a voxel array in a GPU while a more efficient method, using octrees, has been reported [21]. Similar systems have employed octree methods [[28], [29]]. Ref. [22] reports a volumetric mapping application that was found to operate faster on a CPU using octrees than using voxel arrays on a GPU.

3) **Solid-Angle Octrees.** The use of hierarchical tree structures, octrees, to represent volumetric regions of space was extended to represent light and light transport. In a new data structure, the solid-angle octree ("SAO"), nodes represent hierarchical subdivisions of direction space using solid-angle elements ("saels"). Thus, SAOs represent directions in space rather than the space itself. They model all of the rays within volumetric regions projecting out from (or projecting into) a point in space, the center of the SAO's universe. This point is typically attached to a volumetric region and the light entering or exiting the region is represented by light entering or exiting the point. Often two SAOs are attached to a point. Entering light is represented by an "incident" SAO and exiting light by an

"exitant" SAO. Saels are typically projected onto projection planes attached to octree nodes representing mediels. Then an exitant SAO for the point is computed from its incident SAO using the mediel's BLIF. To avoid confusion, an octree containing mediels is referred to as a VLO.

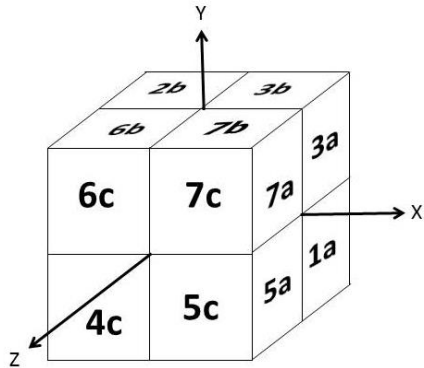

Fig. 5. SAO Face Labeling

The root of the SAO, at level 0, represents all rays emanating from or entering the center point. Below this, each sael is the space occupied by an infinite pyramid truncated by one of the six faces of the SAO's universe. At level 1 in a conventional octree there are eight child nodes. Of the six faces of these nodes, three have external faces as shown in Fig. 5. They are perpendicular to the x, y and z axes and are labeled a, b and c, respectively. There are equivalent a, b and c faces on the negative axes (hidden in the diagram). The volumetric space from the center point through the 24 level 1 faces (8 nodes x 3 faces) are the 24 "top saels." They are handled separately because they have different VLO traversal sequences and may project on to different projection planes.

Fig. 6 is a 2D view of several saels in an SAO. Sael A is a top sael that spans 45° in the x-y plane (a few of the rays enclosed are shown for illustration). A is defined by two planes (rays in 2D), represented by where they intersect the face of the universe in x (face 7a in this case). This is the "span" in y on the face. In 3D there is a similar span in z for this sael. The end points of the span are the points "u" for "upper" and 'l" for "lower" in the diagram. In 3D there would, of course, be two points for each span endpoint in y, at different z values (z = 1 and z = 0 in this case). As described below, spans on projection planes parallel to the sael's face become important when a sael intersects a VLO node containing media.

Saels are projected in only one direction, the positive direction from the center toward its associated face. A negative sael is the "antipode" of a sael, the exact opposite on the other side of the center point. In the diagram, sael "-A" (intersecting face 0a) is the antipodal sael of A. The projection plane will move around as the tree structures are traversed. If it is in the antipode of a sael, the lower value is greater than the upper value (l' and u' in the diagram).

Sael B is the result of the subdivision of a level 1 sael (not shown) and is thus at level 2. Its span is half the length of the parent sael (one-quarter of the parent area in 3D). Likewise, sael D is at level 3. Note that the saels are divided by two with each subdivision but the projected angles are not. The illumination values attached to SAOs are typically weighted to account for the differences in solid-angle area.

**4) Plenoptic Octrees.** A "plenoptic" octree is a 5D spatial data structure that consists of both a conventional octree (or multiple octrees combined with set operations) representing space with mediels, plus SAOs representing sets of directions at various locations in space. The purpose of the new structure is to facilitate the interaction of light and media in a scene by simplifying the projection mathematics and to enable the spread of the calculations over large numbers of tiny processors.

A simplified plenoptic octree is shown in Fig. 7 (in 2D). The upper node on the right is a VLO mediel node representing (with its attached properties) a material that interacts with light. The other node shown is the center of an SAO. A typical task is to project the light in the saels of an exitant SAO out into space and to transfer its illumination to the VLO node, where it will be used as part of the node's incident light (represented with an incident SAO). Then, in turn, it is used to compute its exitant illumination SAO based on the mediel's BLIF. It can also be used, for example, to project sampled illumination from an image (incident light) back on to its source (VLO nodes) where it will contribute to attached exitant SAOs.

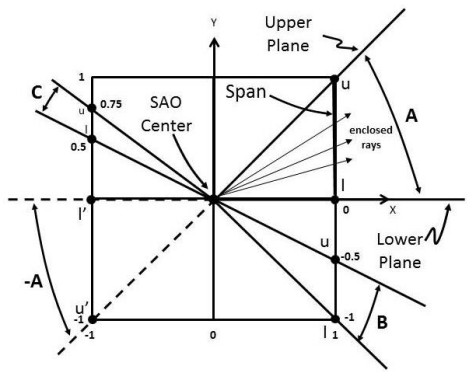

Fig. 6. Sales in SAO (2D)

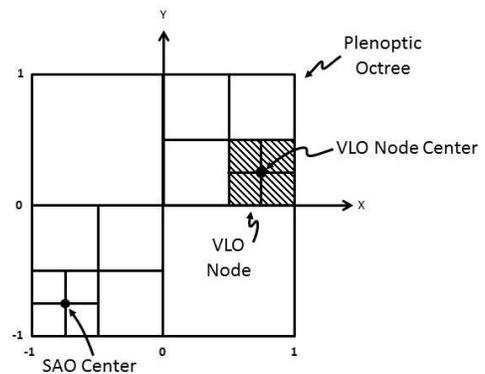

Fig. 7. Plenoptic Octree

5) **Advantages.** Octrees, SAOs and plenoptic octrees have the following characteristics that are of major importance in scene modeling and reconstruction:

- Volumetric – Elements are at an efficient, high level of abstraction. A sael, for example, contains an infinite number of rays that are processed simultaneously.

- Spatially-sorted – They are sorted in space (direction space for SAOs) and can be accessed and traversed spatially. This brings the efficiency normally associated with databases (e.g., handling unlimited data) to the processing of spatial data.

- Multi-resolution – Matter and light are represented at multiple resolutions (space and solid angle, respectively) within the same model (higher resolution at lower levels of the trees). This facilitates successive refinement, such as higher resolution for closer objects in projective space.

- Hierarchical – The material and light datasets represent the entire model at each level of resolution. Parent nodes typically contain a representation of the properties of their child nodes (e.g. average, min/max, count).

6) **Algorithms.** The characteristics of the representations support advanced algorithms. For example, coarse-to-fine algorithms can be directly implemented. Thus, processing can move seamlessly in resolution space (up and down in the tree structures) on-the-fly within one unified data structure. This is performed as needed, based on the immediate needs and results found. Typically success criteria are computed during tree traversal and compared to some set of requirements specified for that region of space (e.g., object of interest to be reconstructed). For example, computations can be terminated in a region if a sufficient level of precision has been obtained.

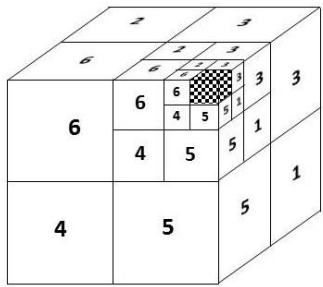

Fig. 8. FTB Sequence

Another advantage is in directional search. Octree nodes can be easily visited in a spatial direction. As illustrated in Fig. 8, a recursive application of a tree traversal sequence of 7 to 0, as shown, will visit nodes in a front-to-back (FTB) traversal where hidden nodes (from the viewing octant shown) can be skipped over.

This efficiency mechanism is typically implemented through the use of a "shadow" quadtree (2D version of an octree) or a q-z tree [13]. It is set up in alignment with the display screen to maintain a map of opaque node projections as nodes are visited in a FTB sequence [15]. This eliminates the traversal of any subtree blocked by something visited earlier that blocks the light. Such hidden nodes (from the current projection point) are simply ignored (not even accessed or generated). In addition, octree traversal terminates when the projection of a subtree falls below some specified size. Often the vast bulk of high-resolution data is simply ignored. This use of shadow quadtrees has been employed for some time in specialized commercial 3D image generators. This is extended in the SPU to model light flow efficiently. It is used to find the nearest material that a sael intersects as it projects out into space from the center of the SAO. If opaque, the matter beyond is not accessed or processed.

This is used when projecting SAO nodes in a plenoptic octree on to VLO nodes. Fig. 9 is an expanded view of the two nodes in Fig. 7. On the left is a top sael after its center has been moved to its location (by traversing the octree). In this case, after the SAO has been located, the VLO is traversed in a FTB sequence to find the first node encountered that modifies the light in some way. The sael is projected on to the current projection plane which is the plane perpendicular to the x axis through the center of the VLO node as shown (other saels may use projection planes through the VLO node center that are perpendicular to other axes).

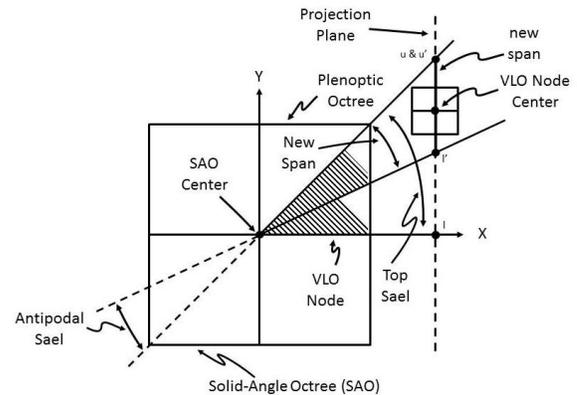

Fig. 9. SAO Projection on VLO Node

At initialization the projection begins at the root of the plenoptic octree for each top sael. Considering just the one shown (in x-y plane in the +x and +y directions), its projection on to the projection plane is first initialized. The starting projection is a plane through the center of the current octree node (the universe), perpendicular to the x axis. In addition, the slopes of the upper and lower planes (in x-y) are initialized to 1,0 and 0,0 respectively.

The projection on this projection plane is maintained as the plenoptic octree is being traversed to locate the center of the SAO. This is simplified by the fact that the movement of the span's end points (in the y direction for the case shown) starts at half the starting slope plus the y movement of the center. It then continues to divide-by-2 with each PUSH. This can be accomplished by simple integer shift/add operations for each PUSH, typically in a single clock cycle in a DSP or simple processing element.

A particular SAO is only defined for a single point in space and represents solid-area directions for a volumetric region around the point. They cannot, in general, be moved to a new location but can be used in generating an SAO for a larger region around it, usually the parent of a VLO node that it is attached to. By traversing the tree to the center point, volumes of decreasing size are visited. SAOs may be attached to each. They may be examined to determine if the traversal needs to continue to higher-resolution SAOs. Also, the SAO center is not required to be at the center of a node. It can be located at another point within the space by modifying the span end points.

Once the SAO is located in space, VLO nodes are then traversed in a FTB sequence, from the root, in a search to find the first VLO node that interacts in some way with the sael's illumination. The projection plane is attached to the VLO nodes and moves with them as they are traversed. The span's end points simultaneously move on the projection plane to maintain the sael projection. As with the sael's movement to its location earlier, the span's end points can be computed with just shift and add operations.

When an interacting VLO node is encountered, the sael may be subdivided into four child saels (2 in the 2D diagram) to bring the projection closer to the size of the node. As shown, a new intersection point (two points in 3D with different z values) is computed half-way between the u and l points. If the PUSH is to the upper sael, as shown, it becomes the new lower bound, l' (the upper point remains the same). Otherwise it becomes the new upper point with the lower point remaining the same.

The child sael is chosen to keep the projection on the center of the VLO node. As before, this can be performed with shift and add operations. This subdivision process usually continues until the area of the projection of the sael on the projection plane is approximately equal to the area of the node itself on the projection plane. In use, an "overlay" of multiple saels is typically used to more precisely measure the projected overlap (and the illumination transfer), as described in [13]. Also, SAO movement and VLO traversal can often be performed simultaneously in the same clock cycle, rather than sequentially.

In most cases, the direction of the illumination projected on to the VLO node is as important as the amount of illumination (e.g., for BLIF calculations). Thus, the illumination is typically transferred to a second SAO attached to the VLO node. It may already exist as a result of earlier projections or be generated when first needed. As shown in Fig. 10, the new sael is the antipodal sael to the original.

In a particular situation either the SAO saels or the VLO nodes can be subdivided to achieve a local goal (e.g., spatial,

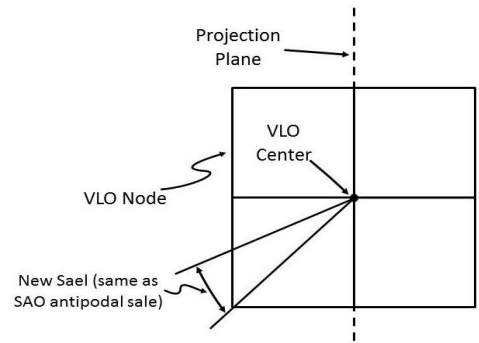

Fig. 10. Sael in VLO

angular and illumination accuracy). If needed, SAOs with centers located to a greater precision may be generated and accessed at lower plenoptic octree levels. After each traversal step, a decision is made to either continue the subdivision process or to stop and process the current nodes (e.g., perform a transfer of illumination). In advanced implementations the subdivision process can be suspended and restarted at a later time if new information becomes available or better results are needed. This is a typical part of bi-directional light-transport processing.

A number of factors contribute to the determination of the next action to be taken. Global accuracy goals are typically combined with local goals and node property information. This typically involves the gradient of the illumination (in the sael), the surface gradient (in the VLO), the directional gradient of the BLIF and so on. Typically, the sael or the VLO (or both) can be pushed to higher levels of resolution for improved accuracy.

**7) Implementation.** To process large amounts of data in a short period of time, algorithms often take advantage of the low-cost of hardware by dividing the computations into multiple computing units operating in parallel. For example, popular algorithms used for processing images and generating images in video games are typically implemented in a Single Instruction, Multiple Data (SIMD) mode of operation. Separate hardware units apply the same sequence of operations simultaneously to data elements from different subsets of the dataset. The success of this approach has resulted in the widespread use of GPUs, typically utilizing a few thousand complex processors, each composed of pipelines of individual computing elements.

This can be efficient if the operations are applied to every element in the dataset. When, however, the algorithm takes advantage of mechanisms that eliminate the need to process every data element, this can be very inefficient. The searching of an indexed database is a common example. The next data element examined is determined by the results of the last computation.

This is a problem when such algorithms are implemented in a pipelined architecture. For example, if the output of the pipeline causes a change in the data to be next entered at the input, the pipeline must be flushed, wasting effort and causing a delay. Such algorithms often exhibit superior performance and reduced power consumption when implemented in CPUs rather than GPUs.

The methods pioneered here are tree-traversal operations where next-step decisions can be made very quickly (often in one clock cycle with a custom implementation). Also, they are inherently parallel operations that spawn independent subtree operations that can be implemented in parallel on large numbers of simple processing elements.

The results of the vehicle hood testing reported below were generated with the SPU implemented in software on a CPU. A hardware implementation using Field-Programmable Gate Array (FPGA) chips is under development.

### III. EXPERIMENTS

We performed two experiments to demonstrate the utility of our GSR approach in reconstructing the shape of non-Lambertian surfaces. In the first experiment, we reconstructed a black surface and a white surface, both containing shallow artificial dents. The resulting reconstructions compare favorably to reconstructions performed by a state-of-the-art optical 3D digitizer. In the second experiment, we reconstructed several automobile panels containing natural hail dents. The results generally agree with the dent locations and sizes as assessed by trained human inspectors using professional inspection equipment. Our results are early examples intended to show the scene reconstruction community what is achievable using our approach. The presented results lead into our efforts, currently underway, on a progression of reconstruction goals on important object types and scene arrangements. These include the following scene characteristics: surface concavity, high self-occlusion, multiple media types imaged together, metal, glass, cast shadows, bright reflections of light sources, moving objects, and an entire room of objects imaged in one session.

#### A. Dent Panel Accuracy Assessment

In this experiment, we manually introduced 16 dents into a thin metal panel in roughly a 4x4 grid arrangement, as shown in Fig. 11.

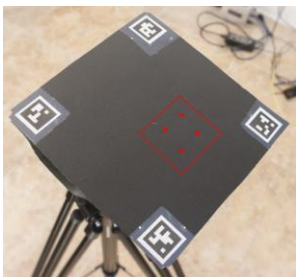

Fig. 11. Black dent panel. This black-painted panel with artificial dents was used to assess GSR reconstruction accuracy. The 2x2 region of deepest dents is outlined in red along with the approximate dent centers.

After anti-glare spray powder was applied, the panel was scanned and reconstructed by a metrology-grade optical 3D digitizer, the GOM ATOS Triple Scan III. We stored the ATOS-generated reconstruction for use as a reference ("ground truth") against which our subsequent GSR reconstructions were compared.

After completion of the ATOS scan, we removed the anti-glare powder from the dent panel and applied 3 thin coats of black spray paint. This was done to realize a BLIF with low diffuse reflectivity (< 1%) to complement the high diffuse reflectivity of a subsequently imaged white panel. We mounted the black dent panel on a tripod and imaged it from 12 inward-facing "object of interest" (OOI) camera viewpoints using a Photon-X PX 3200-R imaging polarimeter at a mean distance of roughly 0.5m from the center of the dent panel. Fig. 5(c) shows a subset of these OOI images. In addition, we imaged the surrounding scene from a multitude of outward-facing "light field of interest" (LOI) camera viewpoints to acquire omnidirectional light field information in the vicinity of the dent panel. 86 images were recorded with the camera pointing in different directions to support reconstruction of the entire hemisphere of incident light at voxels on the dent panel. Fig. 12 shows representative images from the recorded LOI image set.

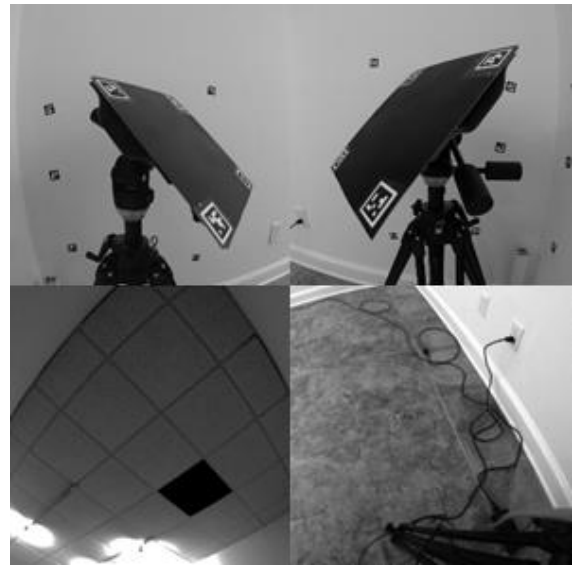

Fig. 12. Examples of reconstruction input images. Top Row: Object-of-interest images. Inward-facing images of the black dent panel (2 of 12 total images used in reconstruction). Bottom Row: Light-field-of-interest images. Outward-facing images of the environment surrounding the dent panel (2 of 86 total images used in reconstruction).

We reconstructed a portion of the panel using a C++/MATLAB implementation of our GSR approach. To limit total processing time in this early, unoptimized implementation, we isolated and reconstructed the 2x2 dent region of greatest mean dent depth (according to the reference reconstruction). A spatially low-frequency version of the reconstructed surface was subtracted from the detailed reconstruction, yielding a 2D depth map that indicates how deeply each dent surfel deviates from an approximate undented (nominal) surface.

Fig. 13 shows overlaid 3D renderings of the depth maps from our reconstruction and the ATOS reference reconstruction. A simple global alignment was performed by subtracting each reconstruction's mean depth (Z coordinate value) from the surfel depths. The RMSD (root-mean-square deviation) between the two reconstructions, taken over all surfels in the region, is approximately 21μm (microns). A cross section through one of the 4 reconstructed dents is plotted in Fig. 14.

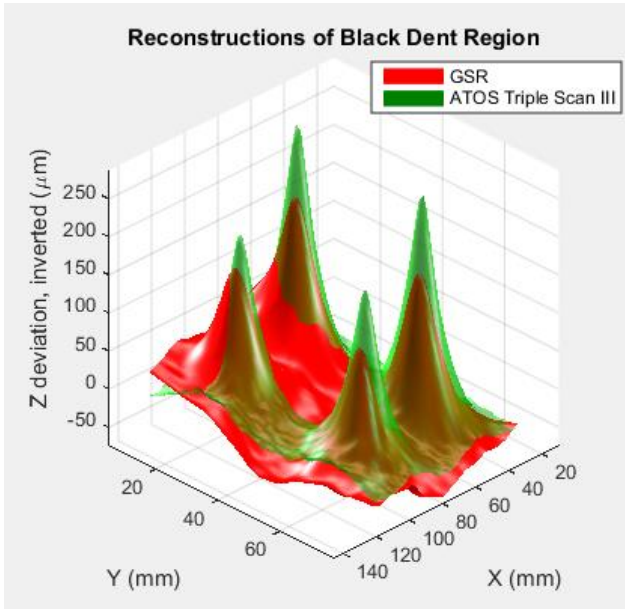

Fig. 13. 3D comparison of black dent reconstructions. A GSR reconstruction (red) and an ATOS Triple Scan III reference reconstruction (green) were performed on the black dent panel.

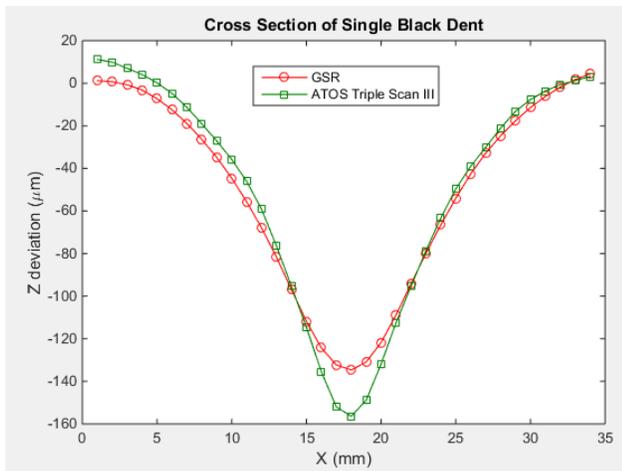

Fig. 14. Cross section comparison of black dent reconstructions. A GSR reconstruction (red) and an ATOS Triple Scan III reference reconstruction (green) were performed on the black dent panel. The cross section is taken across one of the 4 dents shown in Fig. 13

The RMSD in depth over the cross section surfels is approximately 8μm. In this early result, our GSR reconstruction was thus found to be roughly equivalent to that produced a state-of-the-art optical 3D digitizer.

Fig. 15 shows that after imaging and reconstructing the black-painted dent region, we applied 3 thin coats of white spray paint to the same 2x2 dent region. This was done to realize a BLIF with much higher diffuse reflectivity ($> 20\%$) than for the black-painted case above, thus providing a data point on reconstruction performance when a key reflectance characteristic takes on two very different values. The white surface reconstruction scenario is especially salient because materials lighter in appearance tend to polarize more weakly than those darker in appearance.

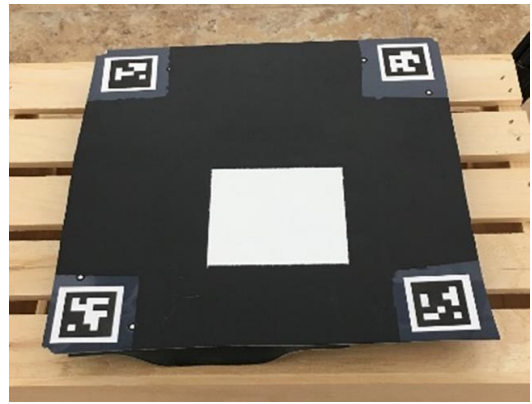

Fig. 15. White dent panel. The black dent panel's 2x2 region of deepest dents was painted white.

Our imaging and reconstruction process for the white panel region was similar in all key respects to that of the previous black region. As seen in Fig. 16, the GSR reconstruction of the white region compares favorably to the reference reconstruction.

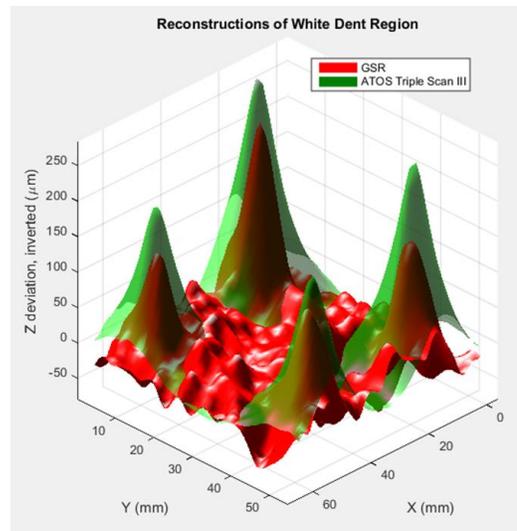

Fig. 16. 3D comparison of white dent reconstructions. A GSR reconstruction (red) and an ATOS Triple Scan III reference reconstruction (green) were performed on the white dent panel.

The RMSD in depth versus the ATOS reconstruction is approximately 45µm. We expect improved accuracy on all types of surface material as we continue to refine our light transport modeling and camera calibrations.

The reconstruction accuracy of these results may also be stated in relative terms as (better than) "one part in a thousand" because the volumetric region containing the 2x2 dent region extends roughly 50mm in X, Y, and Z. Dividing the absolute RMSD by the linear extent of the reconstructed region yields a number indicating "relative error", "parts in a thousand accuracy", and so on. The above results are summarized in Table I.

TABLE I. BLACK AND WHITE DENT PANEL RESULTS

| Imaged Surface | Reconstruction Quantity | | | |
|---|---|---|---|---|
| | *Mean diffuse reflectivity* | *Mean degree of linear polarization* | *Absolute depth error (µm)* | *Relative depth error (parts/thousand)* |
| Black 2x2 dent region | 0.5% | 0.50 | 21 | 0.4 |
| White 2x2 dent region | 22% | 0.03 | 45 | 0.9 |

### B. Automotive Hail Damage

In this experiment, we imaged and reconstructed 17 shiny vehicle panels (hoods) containing naturally caused hail dents of varying diameter and depth. 4 of the panels with different reflective characteristics are shown in Fig. 17.

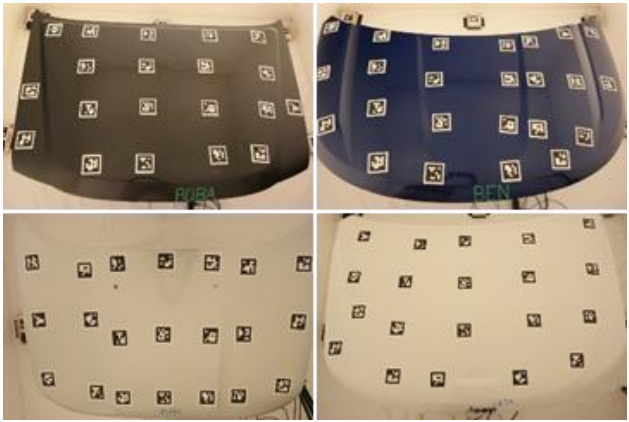

Fig. 17. Imaged automobile panels containing natural hail dents. Coded optical targets are used in regularizing the shape of each reconstructed panel.

The panels were staged inside an enclosure where light from several banks of LEDs filters through diffusing fabric on the four walls and the ceiling. The main purpose of the lit enclosure is to provide sufficient illumination to allow short exposure times that decrease the total imaging time per panel. Despite the diffusing fabric, the interior light field remains anisotropic enough that real light field imaging and reconstruction is required. In other words, an assumed perfectly isotropic light field fails to represent the real light field faithfully enough to yield accurate reconstructions of the shiny vehicle surfaces in this experiment. The imaging setup is shown in Fig. 18.

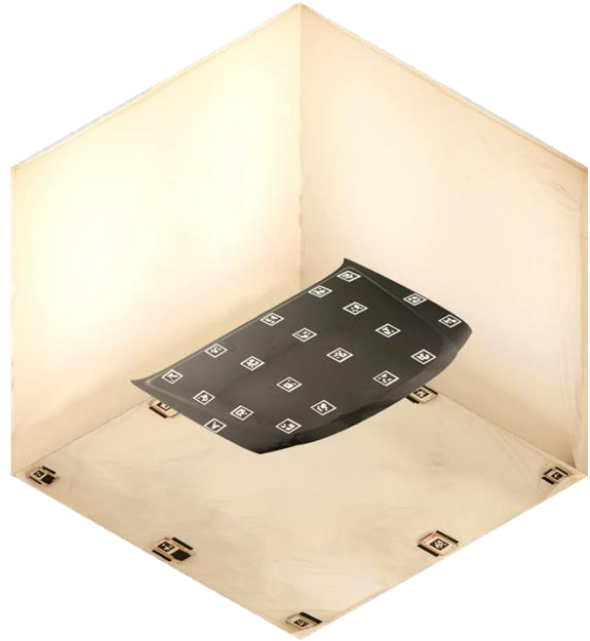

Fig. 18. Imaging setup for automobile panels. The lit enclosure yields a bright light field that allows pragmatically short exposure times when imaging black panels.automobile panels containing natural hail dents.

In addition to GSR imaging, each panel also underwent dent annotation by trained human inspectors. By observing reflections of contrast boundaries from multiple viewpoints (deflectometry principle), the inspectors placed an annotation sticker at the perceived center of each dent found. The sticker color indicates the standard dent size category as used in the hail damage assessment industry. Fig. 19 shows one of our imaged panels with annotation stickers in place. Fig. 20 shows the false-color depth map of a representative region alongside the annotated region on the physical panel.

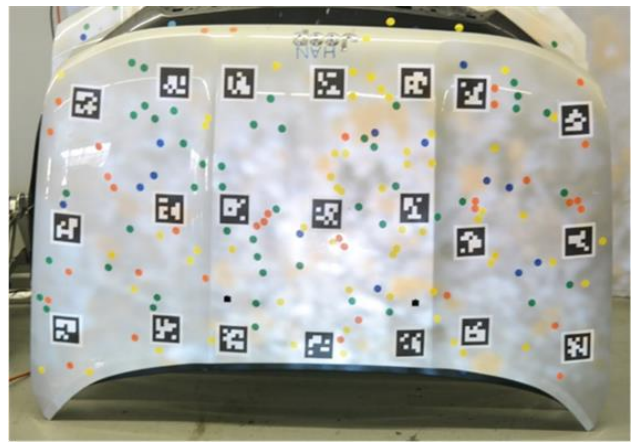

Fig. 19. Physically annotated automobile panel. Trained human inspectors applied an annotation sticker to each discovered dent. A sticker's color represents the dent size category as gauged by the inspector. (Projected pattern to aid ancillary photogrammetric reconstruction is also visible.)

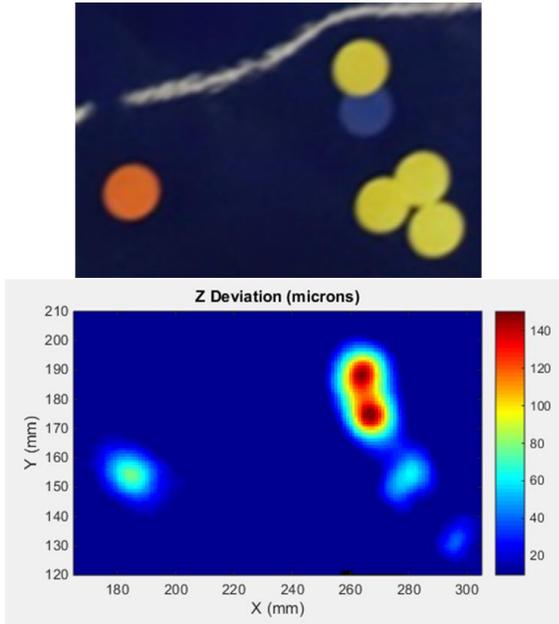

Fig. 20. Reconstruction of dented region of automobile panel. Top Row: Dented region of the panel with size-colored annotation stickers applied by human dent inspectors. Bottom Row: False-color depth map of the GSR reconstruction of the dented region.

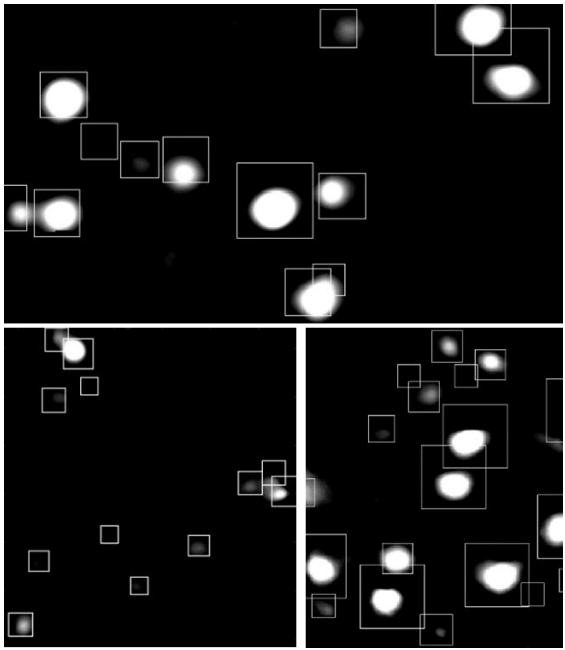

Fig. 21. Annotated dent region depth maps. These are grayscale depth maps of representative reconstructed regions from (Top) black, (Bottom Left) blue, and (Bottom Right) white automobile panels. The overlaid rectangles indicate dent size as estimated by professional human inspection of the physical panels..

We performed a sparse 3D reconstruction of each sticker-annotated panel using commercial photogrammetry software. Our GSR implementation used the coded optical targets (large square labels) in the photogrammetric reconstruction to bring the centers of the annotation stickers into the coordinate system of our reconstruction. Fig. 21 shows rectangles, sized according to the annotated dent size categories, overlaying grayscale depth maps of 3 representative panel regions. Table II presents a basic comparison between our reconstruction and the human inspectors' annotation for each of the 3 regions.

TABLE II. DENTED VEHICLE PANEL RESULTS

| Panel Color | Reconstruction Quantity | | | |
|---|---|---|---|---|
| | Total dents > 20μm in GSR depth map | GSR dents intersecting an annotation rectangle | Annotation rectangles not intersecting a GSR dent | Total dents found by inspectors |
| Black | 11 | 11 | 1 | 13 |
| Blue | 12 | 11 | 1 | 12 |
| White | 15 | 15 | 4 | 19 |

## IV. CONCLUSION

In conclusion, we presented a new approach to scene reconstruction that applies to generalized scenes. This goes beyond the current state of art in the field. We introduce several new innovations in this context: i) a new hierarchical, spatio-directionally sorted data structure called a plenoptic octree to represent the light/matter field that exists in a generalized scene; ii) a new projection method to efficiently compute light transport in a generalized scene; and iii) we extend/blend the concepts of multi-view stereo and shape-from-X, including shape-from-polarization, to GSR in ways that include: using Bidirectional Light Interaction Functions (BLIFs) to represent the potentially omnidirectional interaction of light with homogeneous and non-homogeneous media, and using potentially disjoint surface elements (surfels) to separate regions within a media element (mediel) in the matter field.

We described our early implementation of the GSR approach embodied in C++ and MATLAB software in conjunction with an imaging polarimeter. Experimental results are presented for reconstructions performed on two types of real object surface: a dented test panel and several dented automobile panels. In the case of the test panel, the GSR reconstruction compares favorably to a reconstruction performed by a state-of-the-art optical 3D digitizer. We listed the characteristics of increasingly challenging scenes that form a progression of follow-on reconstruction experiments.

In looking forward, we will continue advancing software and hardware aspects of our GSR implementation. Software improvements fall in two categories: physics modeling of light transport, and scene model optimization. For a generalized scene including objects beyond the automobile panels we experimentally reconstructed, for example, a partially transparent flower vase, we will upgrade our light transport model to account for several transmissions and reflections of each radiometric light field element. In the scene learning module, we would like to efficiently perform an optimization over the full multidimensional degrees of freedom of the matter field and the light field. The main challenge in this regard is to solve for the global minimum of the BLIF-based cost function. Recently, genetic algorithm [[23], [24]] and machine learning techniques [[25], [26]] have been used in bundle adjustment-type nonlinear optimizations. Reinforcement learning [26], in particular, is a possible candidate for efficient solution of our scene model optimization problem as the modeled degrees of freedom increase to accommodate more complex scenes. Our planned hardware improvements involve the acceleration of spatial processing functions performed on plenoptic octrees. We will continue moving toward the use of massively parallel FPGA computing fabrics (as opposed to pipelined GPUs).